\pdfoutput=1

\documentclass[11pt]{article}

\usepackage[]{ACL2023}

\usepackage{times}
\usepackage{latexsym}

\usepackage[T1]{fontenc}

\usepackage[utf8]{inputenc}

\usepackage{microtype}

\usepackage{subfigure}
\usepackage{graphicx}
\usepackage{amsthm}
\usepackage{amsmath}
\usepackage{bm}
\usepackage{bbding}

\usepackage[T1]{fontenc}

\usepackage[utf8]{inputenc}

\usepackage{microtype}

\usepackage{inconsolata}

\usepackage{amsfonts,amssymb}
\usepackage{float}
\usepackage{algorithm}
\usepackage{algorithmic}
\usepackage[normalem]{ulem}

\usepackage{xcolor}

\usepackage{multicol}
\usepackage{multirow}

\title{Personalized LLM Response Generation with Parameterized User Memory Injection}

\author{Kai Zhang$^{1}$, Yejin Kim$^{2}$, Xiaozhong Liu$^{1}$\Thanks{Corresponding Author} \\ $^{1}$Worcester Polytechnic Institute, Worcester, USA \\ $^{2}$The University of Texas Health Science Center at Houston, Houston, USA \\
\texttt{\{kzhang8, xliu14\}@wpi.edu, Yejin.Kim@uth.tmc.edu} \\}

\begin{document}
\maketitle

\begin{abstract}
    Large Language Models (LLMs) have exhibited remarkable proficiency in comprehending and generating natural language. On the other hand, personalized LLM response generation holds the potential to offer substantial benefits for individuals. However, existing work struggles with efficiently incorporating user information for LLM personalization. In this study, we draw inspirations from real-world bionic memory mechanism to propose a novel parameterized \textbf{M}emory-\textbf{i}njected approach using parameter-efficient fine-tuning (PEFT), combined with a Bayesian Optimisation searching strategy to achieve \textbf{L}LM \textbf{P}ersonalization(\textbf{MiLP}). Our MiLP takes advantage from the alignment between real-world memory mechanism and the LLM's architecture. Extensive experiments have shown the superiority and effectiveness of MiLP. To encourage further research into this area, we are releasing our implementations\footnote{https://github.com/MatthewKKai/MiLP}.
\end{abstract}

\section{Introduction}
\label{sec:intro}
The undeniable capability of large language models in comprehending and producing natural language has been underscored by various studies \citep{brown2020language, chowdhery2022palm, touvron2023llama}. Simultaneously, there exists untapped potential to customize these models for delivering personalized responses to users, enabling them to receive tailored and fitting replies according to their individual requirements \citep{bender-koller-2020-climbing}. For instance, in an LLM-based medical dialogue scenario, an assistant capable of recognizing the patient's medical history can generate more tailored responses, rather than offering generic and potentially inappropriate suggestions\citep{huang2023memory}. Individuals in regions with limited access to the medical resources can benefit significantly from such applications, highlighting the imperative needs for personalized LLM response generation \citep{chen2023contributions}. \par

\begin{figure}
    \centering
    \includegraphics[width=7.7cm]{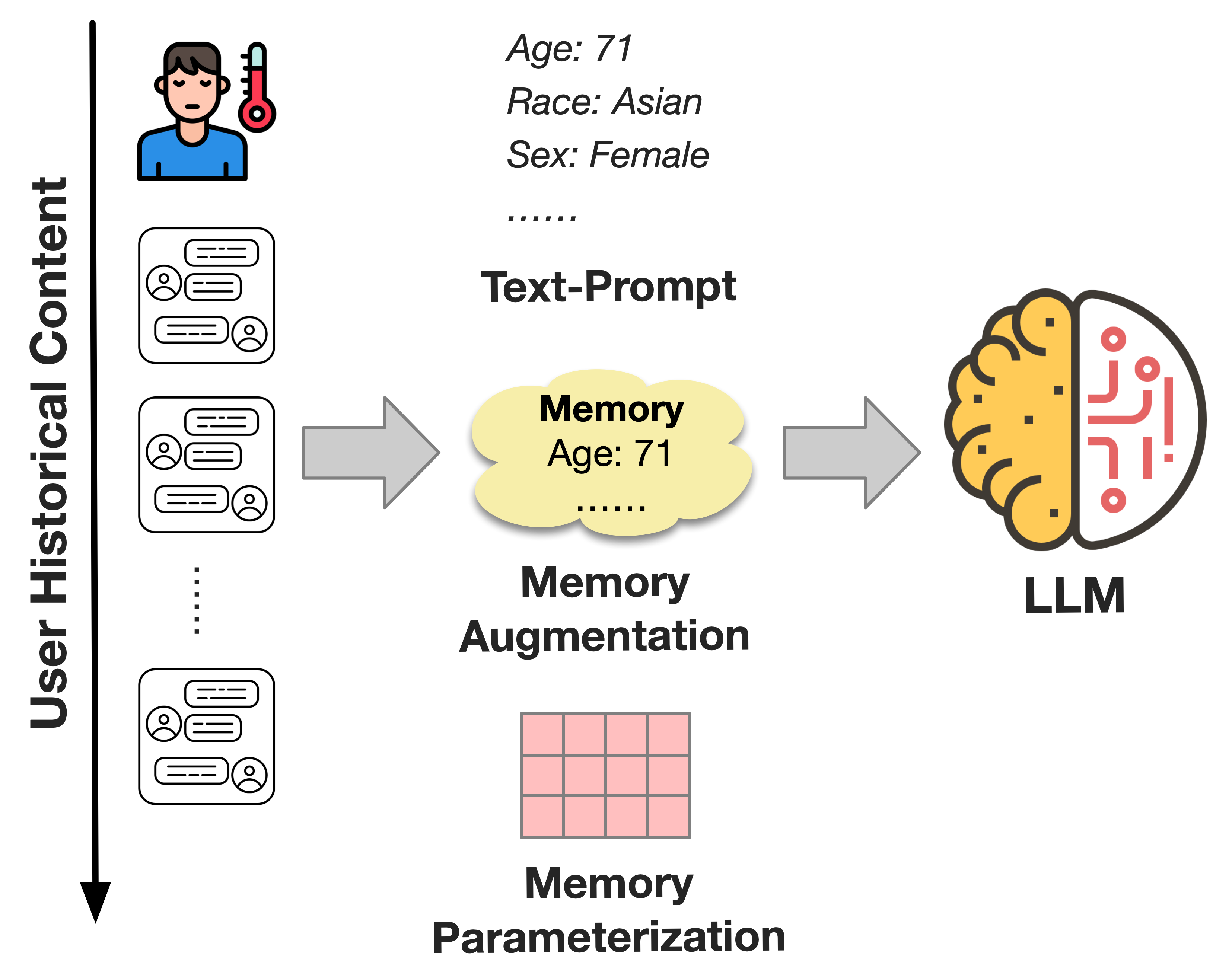}
    \caption{Three types of methods for incorporating user historical content to achieve personalized LLM.}
    \vspace{-10pt}
    \label{fig:toy_example}
\end{figure}

Incorporating user historical information properly to LLM can be a key towards LLM personalization. Existing works can be concluded into three lines as illustrated in Fig.\ref{fig:toy_example}. Text-Prompt based methods draw inspiration from in-context learning to organize the user historical content as prompts, providing them to LLM so that personal information can be considered \citep{petrov2023generative, kang2023llms, liu2023chatgpt}. However, this paradigm suffers from the long-context window limitation of LLM \citep{liu2024lost}. Memory-based approaches provide a solution by leveraging a memory to store user historical content. When a new user query comes, a retriever will first retrieve relevant user information from the memory to prompt LLM produce personalized responses \citep{dalvi2022towards, madaan2022memory, lewis2020retrieval, zhang2023memory}. Unfortunately, they are limited in capturing fine-grained information due to the nature of similarity comparison retrieval process \citep{zhang2023memory}. Additionally, user historical content can be complex and noisy, posing difficulties for LLMs to focus on the most relevant information without a proper learnable process. For example, in healthcare applications, LLMs are used to analyze a patient’s long-term medical trajectory in clinical decision support systems. To understand the underlying mechanisms of disease progression and its interactions with comorbidities, LLMs require comprehensive and precise observations in memory, rather than a few potentially similar records (Appendix \ref{appendix:ragVSmilp}). Fragmented or incomplete records retrieved from memory would provide only a limited and even incorrect snapshot of the disease\footnote{https://www.epic.com/epic/post/cool-stuff-now-epic-and-generative-ai/}\citep{liu2024lost, cosentino2024towards}. To address this, recent studies have proposed parameterizing and projecting user historical content into a learnable representation space \citep{ning2024user, deng2022toward, zhong2022less}. Instead of using text to prompt LLMs, the learned user representations can be neglected in the LLM's decoding process to enable personalized response generation. In this study, we take a further step by investigating a memorization process that mimics real-world memory mechanisms to incorporate user information, aiming to achieve personalization while mitigating associated challenges. \par


Previous studies in neuroscience have indicated that memory is stored in different parts of the brain and is activated accordingly when needed \citep{levenson2005epigenetic, nadel2012memory}. Concurrently, efforts have revealed that the Feed Forward Layers (FFL) of Transformer architecture serve as a memory bank, storing both shallow patterns (e.g., sentences ending with a certain word) and semantic patterns (e.g., sentences about a certain topic) \citep{tay2022transformer, geva-etal-2021-transformer, chen2024multi}. Subsequent attempts have been made to inject external knowledge into LLMs via parameter-efficient fine-tuning (PEFT) \citep{houlsby2019parameter, pfeiffer2020mad, li2021prefix, hu2021lora}, maintaining a modular and adaptable structure without compromising the LLM's original capabilities compared to fully fine-tuning \citep{ye2023qilin, wang2020k, diao2023mixture, yao2022kformer, wang2020k}. Drawing valuable inspiration from the alignment between real-world bionic memory mechanisms and LLM's memory mechanisms, we propose to first parameterize user historical content and store it as memory in the LLM via adapters, followed by fine-tuning the LLM for personalized response generation.\par

However, different memories, with different characteristics highlighting the distinct sensitivity to the allocated parameter budget and the location of the injected layers\citep{he2021towards, zhang2023memory}. Unfortunately, most PEFT applications are limited to a single PEFT architecture with fixed decisions on its components (e.g. hidden size, insertion layers) which can not store and activate different memories for personalization. To address this, we propose to leverage multiple PEFT modules (e.g., LoRAs) for different memory storage and utilize a high-dimensional multi-objective Bayesian optimization (BO) approach to determine the optimal configurations for memory storage. In tandem, we draw inspirations from the alignment between real-world bionic memory mechanism and the LLM's memory mechanism to propose a novel parameterized \textbf{M}emory-\textbf{i}njected method that capitalizes on PEFT, complemented by a novel Bayesian Optimization-based searching strategy to handle multi-PEFT settings for achieving \textbf{L}LM \textbf{P}ersonalization (\textbf{MiLP}). Our contributions can be outlined as follows:\par
$\bullet$ In contrast to previous studies, we leverage the alignment between real-world memory mechanisms and the LLM's architecture to inject parameterized user memory directly into the LLM which offers a fresh perspective for the community on the topic of LLM personalization. \par
$\bullet$ To achieve parameterized memory injection, we propose the MiLP framework, which integrates a comprehensive searching space and a Bayesian optimization-based approach to handle multi-PEFT settings for LLM personalization. \par
$\bullet$ Our empirical results demonstrate significant improvements over three datasets compared to three baselines which validates the effectiveness and superiority of our MiLP approach. \par

\section{Methodology}
\begin{figure*}
    \centering
    \includegraphics[width=\textwidth]{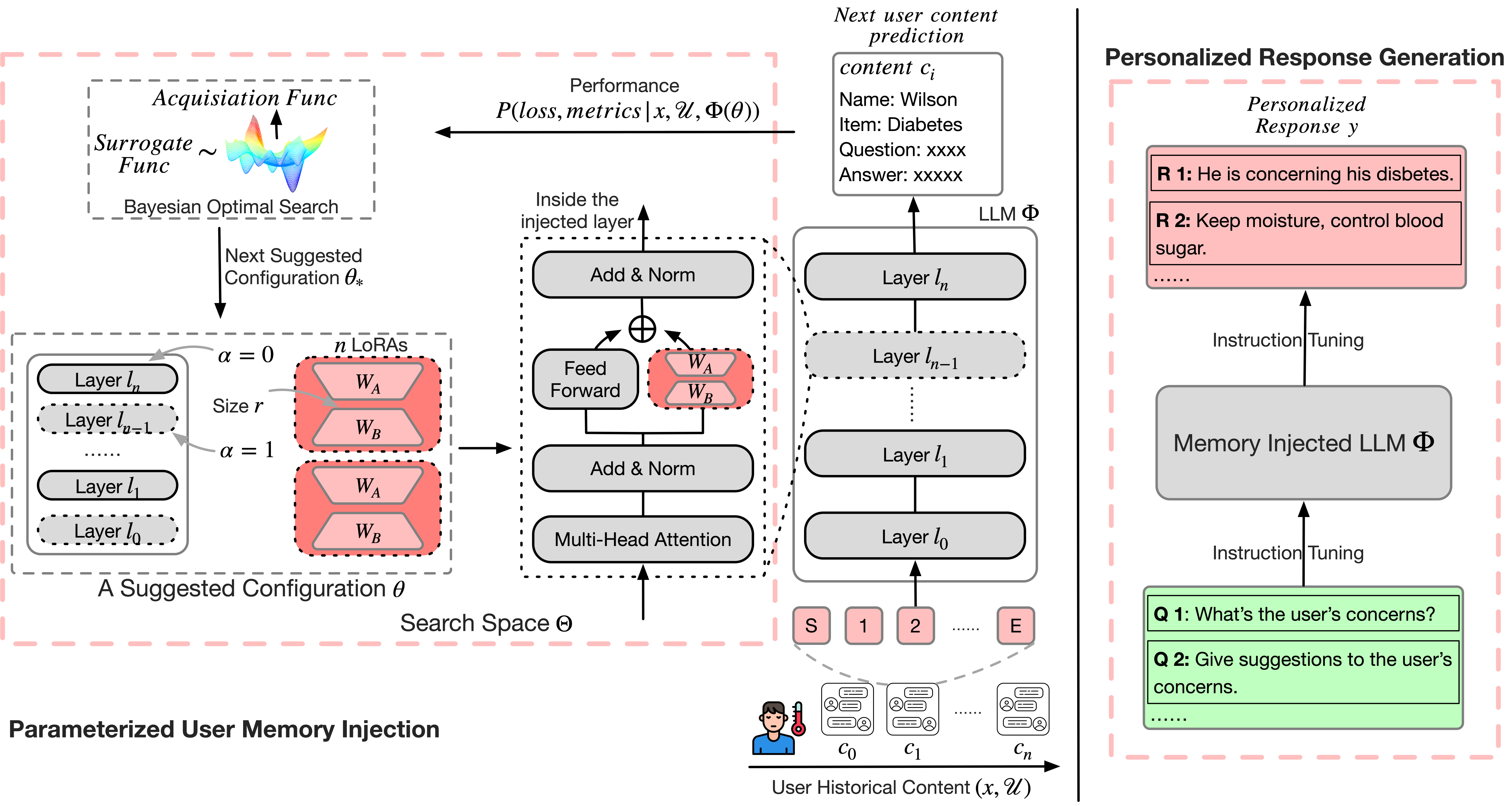}
    \caption{Illustration of the proposed MiLP: The search space encompasses \textit{the number of LoRAs} $n$, \textit{inserted layers} $\alpha$ and \textit{the size of injected LoRA}. Given a suggested configuration $\theta$ (e.g., two LoRAs with fixed size $r$ are injected into the 0-th and (n-1)-th layer, respectively), the base LLM trains on this configuration and take the performance as target. Then the BO search will make a new suggestion and iterate the process until it converges.}
    \label{fig:overview}
\end{figure*}

\textbf{Overview}
Our proposed MiLP takes user's content including user profile, historical content (e.g., dialogues, posts) $\mathcal{U} = \{c_{0}, ..., c_{n}\}$ and a query $x$ as input and the goal is to inject and search for proper memory to output personalized response $y$. The parameterized user memory injection is achieved by applying multiple Low-Rank Adaption (LoRA) modules into the FFL of the base LLM $\Phi$ under an optimal configuration and a modified Bayesian Optimisation approach is utilized to handle the multi-LoRA setting as illustrated in Fig \ref{fig:overview}. The LLM's performance $p$ (e.g., loss, metrics) will be targeted by the optimal search and this process will iterate until it converges. Finally, an instruction-tuning will be performed for aligning the generated response with human intents.  

\subsection{LoRA Module}
Previous work have provided insights in the success of injecting knowledge into the LLM via PEFT tuning \citep{yao2022kformer, wang2020k}. Inspired by the function analysis of feed-forward layers in Transformer\citep{geva-etal-2021-transformer}, our MiLP modified the usage of Low-Rank Adaption (LoRA) \citep{hu2021lora} to the feed-forward layers of the base LLM. For a feed-forward layer $h\ =\ W_{l}x$, the forward process is modified to be:
\begin{equation}
    h = W_{l}x + BAx
    \label{eqLoRA}
\end{equation}
where $W_{l}\in\mathbb{R}^{d\times k}$ denotes the weights of the $l-th$ feed-forward layer, $B\in \mathbb{R}^{d\times r}$, $A\in \mathbb{R}^{r\times k}$ are the low-rank decomposed matrices and the rank $r\ll \min(d, k)$.

\subsection{Parameterized Memory Injection}
Determining how to properly store and activate pertinent information presents a challenge. Inspired by the success of neural architecture search, we start by defining a search space. Subsequently, we employ a Bayesian optimization (BO) approach to identify the optimal configuration for generating personalized responses. In the following sections, we provide a detailed explanation of our search space design, along with the rationale behind it, and describe the process of conducting BO.
\subsubsection{Search Space}
\label{sec:searchSpace}
Search Space plays a pivotal role in searching the optimal configuration for the suitable parameterized memory storgae within LLM. Inspired by \citet{zhou2023autopeft}, we define our seaching space as follow:\\
\textbf{Inserted Layer} Different feed-forward layer within the LLM stores distinct information \citep{geva-etal-2021-transformer} where the shallow layers tend to store shallow patterns (e.g., sentences end with a certain word) while deep layers store semantic patterns (e.g., sentences about a certain topic). Consequently, applying LoRA to all layers can lead to suboptimal results. Thus, we introduce a binary parameter $\alpha$ at each layer $l_{i}$ that controls whether the layer is active (i.e., to be inserted) or inactive.\\
\textbf{Number of LoRAs} Our method is designed for a single and it is intuitive that the volume of distinct user content can vary, resulting in a range of learnable spaces for injecting such content \citep{wang2020k}. To address this, we incorporate the number of LoRAs, denoted as $n$, into our search space. \\
\textbf{Low-Rank Size} Prior studies have demonstrated that the performance of LoRA is greatly influenced by the number of adjustable parameters \citep{chen2022revisiting}. Therefore, it is crucial to dynamically adjust its capacity to align with the demands of the specific task to achieve optimal performance. To address this, we follow \citet{zhou2023autopeft} to include the rank $r$ as a parameter in our search space, which signifies LoRA's capability to store user-specific content in memory.\\

\subsubsection{Bayesian Optimal Search}
While much existing work concentrates on identifying a single PEFT module with the best performance, real-world applications often involve optimizing multiple PEFT models, a scenario that has been rarely explored \citep{zhou2023autopeft}. To address this gap, we opt to employ a modified Bayesian Optimization (BO) approach to determine how different parts of injected memory should be utilized in response to a user's query. \vspace{5pt}\\
\textbf{Bayesian Optimization} leverages two key components:\textbf{1)} A probabilistic \textit{surrogate model} to approximate the objective function using previous observations.\textbf{2)} An \textit{acquisition function} that suggests which point in the search space should be evaluated next. The fundamental principle of Bayesian Optimization (BO) is to iteratively select points for evaluation, striking a balance between exploration (searching different areas) and exploitation (focusing on areas likely to yield the best results). The surrogate model estimates the objective function and its uncertainty, while the acquisition function identifies the most promising points to evaluate. By continuously updating the surrogate model and selecting points expected to improve the objective, BO efficiently explores the space for the optimal solution while minimizing the number of evaluations of the costly objective function. \vspace{5pt}\\
\textbf{Surrogate Function} Applying BO to our defined search space is non-trivial. Thus, we opt for the usage of Sparse AxisAligned Subspace (SAAS-GP) \citep{eriksson2021high} to serve as the surrogate function. SAAS-GP employs robust, sparsity-inducing priors to address the challenge of modeling high-dimensional data. It assumes that despite the nominal high dimensionality, the effective dimensionality is significantly lower, thereby simplifying the modeling process. Given the user's content $\mathcal{U}$, a query $x$ and the base LLM $\Phi$ with a suggested configuration $\theta$, the performance $p(l, rl|x, \mathcal{U}, \Phi(\theta))$ can be represented by the CrossEntropyLoss $l = -\frac{1}{N}\sum\nolimits_{i=1}^{N}logP(y_{i}|y_{<i}, \mathcal{U}, x)$, where $N$ is the the length of targeted length, and ROUGE-L score $rl$ between generated $\hat{y}$ and targeted response $y$. Thus, we can give the surrogate function in our settings:
\begin{equation}
    p(\theta)\sim \mathcal{N}(\mu(\theta), \sigma^{2}(\theta))
\end{equation}
where $\theta\in \Theta$ is a suggested configuration from our defined search space $\Theta$ as elaborated in Section \ref{sec:searchSpace}, $\mu(\theta)$ is the mean and $\sigma^{2}(\theta)$ is the variance. For the kernel function, we use log-Normal distribution as the kernel. Then given a new configuration $\theta_{*}$, the posterior distribution of $p(\theta_{*})$ can be updated as follows:
\begin{equation}
    p(\theta_{*})|\{\theta_{i}, f(\theta_{i})\}_{i=1}^{n}\sim \mathcal{N}(\mu_{*}, \sigma^{2}_{*})
    \label{eq:assa-gp}
\end{equation}
where $n$ is the number of observed points. The mean and variance of the posterior distribution are computed using the Gaussian process regression.\vspace{5pt}\\
\textbf{Acquisition Function} 
For acquisition function, we use the Negative Expected Hypervolume Improvement (NEHVI) \citep{daulton2021parallel} since it quantifies the negative expected improvement in hypervolume when including a new point in the solution set which in nature is suitable for handling multi-objective optimization setting. The function in our setting can be described as:
\begin{equation}
    NEHVI(\theta) = -\mathbb{E}[H(p(\theta^{+})\cup {p(\theta)})-H(p(\theta^{+})]
    \label{eq:acquisitionF}
\end{equation}
where $H(\cdot)$ is the hypervolume function, $p(\theta^{+})$ is a reference point representing the best-known objective values achieved so far and $p(\theta)$ is the predicted function value at $\theta$ calculated by the surrogate function. 

\subsection{Personalized Response Generation}
Upon on the learned user representation from historical content, the LLM can be fine-tuned to generate personalized response. We resort to the usage of instruction tuning which has shown great ability for leading LLM to generate desired response in just a few samples\citep{stiennon2020learning, min2021metaicl, ouyang2022training}. In this work, we fine-tune the memory injected model on instruction-following examples in a supervised manner to aligned the LLM's response with human intents with respect to the user historical content. \\


\section{Experimental Settings}
MiLP is tailored to fine-tune the base LLM to generate personalized responses. To evaluate its effectiveness, we compare our method across three public datasets that contain user historical content. For this evaluation, we utilize four different base LLMs of varying scales (Please check Appendix \ref{appendix:scalability} for detailed scalability justification.).
\subsection{Datasets}
\textbf{AmazonQA/Products}\citep{deng2022toward}
is a public E-commerce dataset of which each data sample contains user's historical posted content, including questions, answers and reviews as well as the corresponding product's description\footnote{https://cseweb.ucsd.edu/~jmcauley/datasets.html}. \\
\textbf{Reddit}\citep{zhong2022less}
is a public dataset collect from social media platforms where a user can post question and respond to other users. Each data sample contains a query, a response and a sequence of this user's dialogue history\footnote{https://github.com/bangbangbang12315/MSP/tree/release/data}.\\
\textbf{MedicalDialogue}\citep{zhang2023memory}
is a medical dialogue dataset derived from open-source medical corpus of which each data sample contains a patient's profile, preference and the historical dialogues between the patient and the doctor\footnote{https://github.com/MatthewKKai/MaLP/tree/main/data}.\\
The detailed comparisons can be seen in Table \ref{tab:dataset_statistics}.For our experiments, we split the dataset in a user-oriented manner and format each user's historical content into a fixed text phrase which allows us to perform next user content prediction task to learn the user's preference.
\begin{table}[]
    \centering
    \small
    \begin{tabular}{cccc}
        \hline
         & AmazonQA & Reddit & MedDia \\
         \hline
         \# User & 46,923 & 46,818 & 60 \\
         \# Samples & 51,936 & 95,881 & 10,920 \\
         \# Len(History) & 30.7 & 72.4 & 182 \\
         Avg. Len(Content) & 23.6 & 22.8 & 27.8 \\
         Avg. Len(Response) & 50.2 & 9.1 & 23.7 \\
         \hline
    \end{tabular}
    \caption{Statistics comparison of the datasets}
    \label{tab:dataset_statistics}
\end{table}

\subsection{Baselines}
We opt to compare our MiLP with three different configurations for LLM Personalization: Text-prompt (TpLP), Memory-Augmented\citep{zhang2023memory} (MaLP), User-embedding (UeLP)\citep{ning2024user} in terms of four LLMs as the base models\footnote{Due to the resources limitation, we are unable to test larger scale LLMs.}: DialoGPT\citep{zhang2019dialogpt}, RoBERTa\citep{liu2019roberta}, LLaMA2-7B and LlaMA2-13B\citep{touvron2023llama}. For a fair comparison, we use the configuration with the best performance as reported in their paper.

\subsection{Evaluation Metrics}
\textbf{Automatic Evaluation}
We resort to the usage of \textbf{ROUGE-1} and \textbf{ROUGE-L} to measure the word overlaps between the generated response and the ground truth. Further, since the goal is to generate personalized response, the \textbf{Persona F1 (P-F1)}\citep{ma2021one} is also used to measure the unigram F1 between the generated response and the user's content (e.g., historical dialogues, profile etc.). \\
\textbf{Human Evaluation}
Automatic evaluation can assure the quality of the generated response with respect to the ground-truth, however, we recognize that human evaluation is needed. Thus, we follow the scoring method of \citet{wang2023chain} and calculate the \textbf{Win Rate}, scoring the generated response and compare the scores between different settings and the standard generation of the Text-prompt based method.

\subsection{Implementation Details}
For implementation details, we leverage the Transformers \citep{wolf-etal-2020-transformers} and Adapters\citep{pfeiffer2020AdapterHub} as the base code and conduct extensive experiments with the DialoGPT, RoBERTa, LlaMA2-7B and LlaMA2-13B. We use the AdamW optimizer\citep{loshchilov2018decoupled, paszke2017automatic} with a learning rate of 5e-4 and also a linear warm-up scheduler initialized with 10\% of the total training steps as warm-up steps and a weight decay of 1e-4 to avoid over-fitting for all the experiments. The batch size per device is set to 8. Further, for all the LLMs, we follow their default settings from the Transformers \citep{wolf-etal-2020-transformers} and add search space factors in the their configurations. We modified the forward logic of injected layers by combing the outputs from both the base model layer and the injected lora module. For BO algorithm implementation, we resorted to the usage of BoTorch \citep{balandat2020botorch} and follow the suggested settings from \citet{zhou2023autopeft} for both surrogate function and acquisition function. For prior distributions, we randomly sample 100 initialisation points for all the experiments. For all datasets used, we split 70\% of them as the training set, 10\% of them as the validation set and 20\% of them as the testing set. For search space, each factor is an integer from different ranges. The details can be seen in Table \ref{tab:search-space}. All the experiments are conducted on a computation node configured with four 80G Tesla A100 GPUs.

\begin{table}[]
    \centering
    \begin{tabular}{cc}
    \hline
        Factor & Range \\
        \hline
        $\alpha$ & [0, 1] \\
        $n$ & [0 $\sim$ 32] \\
        $r$ & [8, 16, 32, 64, 96]\\
        \hline
    \end{tabular}
    \caption{Search range for each factor from the space.}
    \label{tab:search-space}
\end{table}

\section{Experimental Results}
\subsection{Comparative Study}
\begin{table*}[]
    \small
    \centering
    \begin{tabular}{ccccccccccc}
        \hline
         \multirow{2}{*}{Model} & \multirow{2}{*}{Type} & \multicolumn{3}{c}{AmazonQA} & \multicolumn{3}{c}{Reddit} & \multicolumn{3}{c}{MedDia} \\
          & & ROUGE-1 & ROUGE-L & P-F1 & ROUGE-1 & ROUGE-L & P-F1 & ROUGE-1 & ROUGE-L & P-F1 \\
          \hline
         \multirow{4}{*}{DialoGPT} & TpLP & 16.44 & 14.63 & 0.741 & 14.57 & 13.89 & 0.337 & 15.47 & 14.31 & 0.890 \\
         & MaLP & 17.02 & 16.31 & 0.843 & 16.12 & 13.40 & 0.399 & 17.15 & 15.87 & 0.929 \\
         & UeLP & 18.02 & 17.74 & 0.901 & 15.95 & 13.71 & 0.389 & 16.92 & 15.04 & 0.916\\
         & \textbf{MiLP} & \textbf{18.61} & \textbf{17.83} & \textbf{0.925} & \textbf{16.38} & \textbf{14.51} & \textbf{0.409} & \textbf{17.67} & \textbf{15.94} & \textbf{1.072} \\
         \hline
         \multirow{4}{*}{RoBERTa} & TpLP & 17.35 & 15.41 & 0.704 & 13.91 & 12.81 & 0.391 & 14.81 & 13.99 & 0.947 \\
         & MaLP & 18.50 & 15.76 & 0.828 & 14.17 & 13.96 & 0.462 & 17.79 & 16.80 & 1.141 \\
         & UeLP & 18.97 & 16.19 & 0.899 & 15.96 & 14.86 & 0.491 & 16.21 & 14.33 & 0.971 \\
         & \textbf{MiLP} & \textbf{19.73} & \textbf{17.59} & \textbf{0.974} & \textbf{16.83} & \textbf{15.09} & \textbf{0.531} & \textbf{18.96} & \textbf{17.18} & \textbf{1.187} \\
         \hline
         \multirow{4}{*}{LlaMA2-7B} & TpLP & 19.61 & 17.71 & 1.817 & 14.37 & 13.70 & 0.533 & 17.19 & 16.77 & 1.818 \\
         & MaLP & 19.80 & 17.06 & 1.834 & 13.91 & 13.09 & 0.533 & 19.98 & 18.89 & 1.917 \\
         & UeLP & 20.91 & 18.79 & 2.083 & 16.61 & 14.74 & 0.613 & 18.27 & 16.73 & 2.081 \\
         & \textbf{MiLP} & \textbf{21.69} & \textbf{19.96} & \textbf{2.176} & \textbf{18.63} & \textbf{16.81} & \textbf{0.756} & \textbf{20.98} & \textbf{19.73} & \textbf{2.274} \\
         \hline
         \multirow{4}{*}{LlaMA2-13B} & TpLP & 24.91 & 23.36 & 2.107 & 20.87 & 20.19 & 0.678 & 22.77 & 21.32 & 2.009 \\
         & MaLP & 22.61 & 21.29 & 2.061 & 21.18 & 20.78 & 0.671 & 23.77 & 22.69 & 2.250 \\
         & UeLP & 25.02 & 23.74 & 2.089 & 22.03 & 21.80 & 0.704 & 22.18 & 20.88 & 2.131 \\
         & \textbf{MiLP} & \textbf{25.51} & \textbf{24.25} & \textbf{2.283} & \textbf{22.28} & \textbf{21.83} & \textbf{0.864} & \textbf{24.13} & \textbf{22.96} & \textbf{2.337} \\
         \hline
    \end{tabular}
    \caption{Comparative results on different datasets using automatic metrics.}
    \label{tab:comparative_study}
\end{table*}

Table \ref{tab:comparative_study} presents the automatic evaluation comparative results between baselines and our proposed MiLP on three datasets. It is evident that the inclusion of memory improves performance across all baseline models, highlighting the ability of provided personal information to enhance personalized response generation. MaLP, which incorporates long- and short-term memory, outperforms text-prompt based methods, indicating the effectiveness of differentiating stored information. However, our proposed MiLP exhibits superior performance compared to them. MiLP achieves average relative improvements of 4.38\%, 5.05\% and 2.09\% in ROUGE-L scores over all base LLMs against the best baselines on three datasets, respectively. One interesting thing we found is that as the base LLMs goes deeper (e.g., the number of hidden layers is larger), the relative improvements will also increase. For example, LlaMA2-13B equipped with MiLP achieves a relative improvement of 0.44\% in ROUGE-L score while DialoGPT equipped with MiLP only achieves a relative improvement of 1.19\% on MedicalDialogue dataset against the best baseline. We attribute this to the deeper layers learning more semantic features \citep{geva-etal-2021-transformer}. \par
Moreover, the improvements in persona-F1 score confirm that incorporating a memory mechanism allows for the integration of more user-specific information into the generated response, thereby enhancing personalization. However, we observed that MiLP demonstrates better coverage of personalized information compared to the best baselines, achieving average increases of 0.090, 0.088, and 0.117 in persona-F1 score across the three datasets, respectively. This can be attributed to the fact that the approach of retrieving pre-stored memory to augment LLM personalization relies on the quality of retrieval and the LLM's understanding of the retrieved prompts, which may lead to sub-optimal results. In contrast, user-embedding-based method anticipated in the LLM's decoding process leading to a better performance. Additionally, our MiLP injects memory directly into the intricate LLM and achieves a better understanding of the injected information through our proposed BO approach, thereby producing more relevant user-specific information when generating personalized responses. The comprehensive results validate the effectiveness and superiority of our proposed MiLP.

\subsection{Quality Study\footnote{We further provide a Case Study which can be seen in Appendix \ref{appendix:case_study}.}}
\begin{table}[ht]
    \small
    \centering
    \begin{tabular}{ccccc}
        \hline
         Model & Type & AmazonQA & Reddit & MedDia \\
          \hline
         \multirow{4}{*}{DialoGPT} & TpLP & - & - & -  \\
         & MaLP & 57.37 & 51.95 & 69.33  \\
         & UeLP & 63.20 & 60.17 & 75.02 \\
         & \textbf{MiLP} & \textbf{63.97} & \textbf{60.76} & \textbf{75.78}  \\
         \hline
         \multirow{4}{*}{RoBERTa} & TpLP & - & - & -  \\
         & MaLP & 57.91 & 56.39 & 63.83  \\
         & UeLP & 59.99 & 60.11 & 66.75 \\
         & \textbf{MiLP} & \textbf{61.97} & \textbf{60.19} & \textbf{67.63} \\
         \hline
         \multirow{4}{*}{LlaMA2-7B} & TpLP & - & - & -  \\
         & MaLP & 64.74 & 59.67 & 88.93  \\
         & UeLP & 65.91 & \textbf{61.870} & 89.43  \\
         & \textbf{MiLP} & \textbf{66.17} & 59.81 & \textbf{91.83}  \\
         \hline
         \multirow{4}{*}{LlaMA2-13B} & TpLP & - & - & -  \\
         & MaLP & 71.82 & 72.96 & 87.89  \\
         & UeLP & 74.37 & 75.13 &  89.18 \\
         & \textbf{MiLP} & \textbf{75.48} & \textbf{76.61} & \textbf{90.67}  \\
         \hline
    \end{tabular}
    \caption{Quality study results on different datasets using the Win Rate metric.}
    \label{tab:quality_study}
\end{table}
We further conduct quality study to examine the quality of generated responses as illustrated in Table \ref{tab:quality_study}. We observed that leveraging a memory achieves above 50\% win rate for all base LLMs over three datasets. We attribute this to the nature that introducing user-specific information as prompts can enhance LLM response generation in terms of personalization. However, relying solely on memory can lead to misunderstandings by the LLM when generating responses, resulting in suboptimal outcomes. Our MiLP not only utilizes user-specific information from the user's historical content but also leverages the natural language understanding and inference abilities of the LLM itself through our proposed BO method. This enables the LLM to comprehend which information should be considered when generating a response, leading to optimal performance compared to other baselines in most scenarios. However, we also notice that as the base LLM becomes more complicated, its greater natural language understanding and inferring ability are not always accompanied with better performance. For example, the results of LlaMA2-13B are incomparable with LLaMA2-7B on MedicalDialogue dataset. We attribute this disparity to the greater sparsity of user-specific information in the historical user content of the MedicalDialogue dataset compared to the other two datasets. Consequently, the configuration space for LlaMA2-13B is relatively sparser than that of other models, leading to suboptimal performance.
The base LLM size selection with respect to the scale of input information (e.g., user numbers, the memory size etc.) is worthy to be explored in the future. Despite these challenges, the increase in win rate confirms the effectiveness of our proposed MiLP.\par

\textbf{Human validation} To validate the alignment of our automatic scoring schema with human judgments, we adopted the methodology of \citet{wang2023chain} for point-wise evaluation. We hired two master's students to evaluate 50 response pairs, consisting of responses generated by standard settings and MiLP using LLaMA2-13B, along with the corresponding user content for each pair. The students were asked to indicate which response they deemed better by selecting 1 (win), 0 (tie), or -1 (lose) for each pair. Subsequently, we calculated the Pearson Correlation Coefficient (P.C) and the accuracy between human selections and automatic selections. The high P.C of 0.86 and an accuracy of 91\% collectively indicate the feasibility and high confidence of our evaluation method.

\subsection{Ablation Study on Search Space}
\begin{table}[ht]
    \small
    \centering
    \begin{tabular}{cccc}
        \hline
         Space & AmazonQA & Reddit & MedDia \\
          \hline
         Num & 2.011 & 0.604 & 2.027 \\
         Size & 2.017 & 0.601 & 2.034 \\
         Layer & 1.921 & 0.597 & 2.001 \\
         Num+Size & 2.016 & 0.604 & 2.073 \\
         Num+Layer & 2.130 & 0.731 & 2.196 \\
         Size+Layer & 2.195 & 0.767 & 2.197 \\
         \textbf{Num+Size+layer} & \textbf{2.283} & \textbf{0.864} & \textbf{2.337} \\
         \hline
    \end{tabular}
    \caption{Ablation study of using LlaMA2-13B as the base on different search space using Persona-F1 score as the metric.}
    \label{tab:ablation_study_1}
\end{table}
To explore the effectiveness of each factor within the search space, an ablation study is conducted. As can be observed in Table \ref{tab:ablation_study_1}, search number $n$ or size $r$ only achieves similar personal information coverage which we attribute to these two factors are more related to the scale of stored memory. In contrast, the choice of which layer to inject influences how the LLM understands the injected memory. During our experiments, when only searching which layers to inject, as the number of LoRAs and LoRA size become larger, the overall performance will be better.However, performance eventually dropped once $n$ and $r$ reached a threshold. This observation aligns with our understanding that there should be a balance between the size of input information and trainable parameters. Meanwhile, without being aware of what memory should be used when generating responses will lead to a sub-optimal result. These findings verify the necessity and effectiveness of the comprehensive search space.

\subsection{Ablation Study on MiLP Components}
\begin{table}[ht]
    \small
    \centering
    \begin{tabular}{cccc}
        \hline
         Components & ROUGE-1 & ROUGE-L & P-F1 \\
          \hline
         Instruction Tuning & 23.67 & 22.07 & 2.107 \\
         Memory Injection & 22.81 & 19.90 & 2.331 \\
         \textbf{Full MiLP} & \textbf{24.13} & \textbf{22.96} & \textbf{2.337} \\
         \hline
    \end{tabular}
    \caption{Ablation study of using LlaMA2-13B as the base on different components on MedDia dataset.}
    \label{tab:ablation_study_2}
\end{table}
To determine whether the effectiveness of MiLP is due to instruction-tuning or the parameterized memory injection component, we conducted an ablation study. The results are presented in Table \ref{tab:ablation_study_2}. As shown, neither instruction-tuning alone nor parameterized memory injection alone can outperform the fully configured MiLP. Additionally, we observed that ROUGE scores are higher with instruction-tuning, while the Persona-F1 score is higher with parameterized memory injection. We attribute this to instruction-tuning aligning the generated response with human intents, while parameterized memory provides more personal information. With all components incorporated, our MiLP demonstrates the highest effectiveness.



\section{Conclusion \& Future Work}
In a nutshell, we propose a novel frame called MiLP to achieve Memory-injected LLM personalization. MiLP uses LoRA as the base PEFT module and leverages a Bayesian Optimisation based approach to iterative inject and search user historical information towards personalized response generation from our defined search space. Additionally, we conduct extensive experiments to compare our method with three baselines on three datasets and the results verify the effectiveness and superiority of our MiLP. Further, an ablation study is conducted for validating the the necessity of each factor within the defined search space.  \par
In the future, exploring scalability with a larger number of users and larger LLMs is essential. Additionally, enhancing the inference ability to better understand user-specific needs is crucial. This includes integrating shared information and user graphs into the LLM to improve personalized response generation.

\section{Related Work}
\textbf{Memory-Augmented LLM} refers to apply a memory that stores extra information for enhancing LLM's output \citep{ouyang2022training}. Various efforts have been made to utilize memory in this context. \citet{tandon2021learning} proposed leveraging a corrector that can rectify the model's output based on similar mistakes stored in memory. However, this method focuses on repairing wrong outputs. In contrast, \citet{madaan2022memory} argued that stored experiences can be used to prevent incorrect outputs by incorporating feedback into the new query. Another usage of memory is to include the memory into a learning frame such as self-learning or teacher-student paradigm so that the LLM can learn by iterative refinement\citep{madaan2023self, dalvi2022towards}. In tandem, the key for better usage of memory is to equip powerful retrievers\citep{guu2020retrieval, lewis2020retrieval, yuan2022selecting} and improve the effectiveness of storing memory \citep{zhang2023memory}. Unlike previous studies, our MiLP framework parameterizes and injects memory directly into the LLM through PEFT modules while accounting for memory budgets. \\
\textbf{Personalized LLM} has garnered increasing attention for its ability to provide tailored experiences that align with user expectations and needs \citep{salemi2023lamp}. Previous works focused on identifying user preferences using Ceteris Paribus (CP)-nets \citep{asher2010extracting} and modeling user historical content into language models \citep{zhong2022less, deng2022toward}. However, these methods suffer from limited natural language understanding ability of language models. With the emergence of LLMs, prompt-based methods have been developed to design detailed prompts that guide LLMs in producing desired outputs while being aware of user status and contextual content \citep{wang2023chain, wu2023tidybot, aher2023using}. Another line of approach attempts to leverages memory to store user relevant information. When a new user query arises, a retriever will retrieve relevant user information from the memory to prompt the LLM to produce personalized responses \citep{dalvi2022towards, madaan2022memory, lewis2020retrieval, zhang2023memory}. Moreover, recent studies have explored projecting user information into embeddings, allowing parameterized user data to participate in the decoding process of the LLM to generate personalized responses \citep{korbak2023pretraining, salemi2023lamp, xu2023baize, ning2024user}. In contrast to previous works, we build on the alignment between real-world bionic memory mechanisms and LLM memory mechanisms. We leverage a novel Bayesian Optimization strategy to inject parameterized user memory into the LLM, enabling it to produce personalized responses.\\
In tandem, our work stands out from previous research as we pioneer a parameterized memory injection method. By leveraging this novel method, user information can be stored and activated effectively to produce personalized responses.

\section*{Limitations}
Despite the empirical success, our approach has two notable limitations that warrant attention. Firstly, our approach relies on the user's historical content, and the sparsity of user-specific information within this content can influence the quality of the generated response. In the near future, recent efficient search strategies on LoRAs, such as LoNAS \citep{munoz2024lonas} and Shears \citep{munoz2024shears}, should be considered for more efficient and robust memory injection. \par
Secondly, our method is designed for a single user. Therefore, it would be valuable to explore how the number of users and the scale of the LLM can impact the generated response (Please check the Appendix \ref{appendix:scalability} for more justifications of scalability.). For example, when dealing with a larger group of users, it would be important to consider how to assign Personalized Fine-Tuning (PEFT) modules (e.g., type, number) for each user and how to select the base LLM (e.g., one single small LLM for one user or one single layer within a large LLM for one user). However, due to computational and data resource limitations, we are unable to explore this at present. We hope to address this in future work and see increasing attention given to this aspect.

\section*{Ethics Statement}
After carefully reviewing the ACL Ethics Policy, we are committed to show our respect and obey to consent all.

\section*{Acknowledgements}

\bibliography{anthology,custom}

\begin{thebibliography}{62}
\expandafter\ifx\csname natexlab\endcsname\relax\def\natexlab#1{#1}\fi

\bibitem[{Aher et~al.(2023)Aher, Arriaga, and Kalai}]{aher2023using}
Gati~V Aher, Rosa~I Arriaga, and Adam~Tauman Kalai. 2023.
\newblock Using large language models to simulate multiple humans and replicate human subject studies.
\newblock In \emph{International Conference on Machine Learning}, pages 337--371. PMLR.

\bibitem[{Asher et~al.(2010)Asher, Bonzon, and Lascarides}]{asher2010extracting}
Nicholas Asher, Elise Bonzon, and Alex Lascarides. 2010.
\newblock Extracting and modelling preferences from dialogue.
\newblock In \emph{International Conference on Information Processing and Management of Uncertainty in Knowledge-Based Systems}, pages 542--553. Springer.

\bibitem[{Balandat et~al.(2020)Balandat, Karrer, Jiang, Daulton, Letham, Wilson, and Bakshy}]{balandat2020botorch}
Maximilian Balandat, Brian Karrer, Daniel~R. Jiang, Samuel Daulton, Benjamin Letham, Andrew~Gordon Wilson, and Eytan Bakshy. 2020.
\newblock \href {http://arxiv.org/abs/1910.06403} {{BoTorch: A Framework for Efficient Monte-Carlo Bayesian Optimization}}.
\newblock In \emph{Advances in Neural Information Processing Systems 33}.

\bibitem[{Bender and Koller(2020)}]{bender-koller-2020-climbing}
Emily~M. Bender and Alexander Koller. 2020.
\newblock \href {https://doi.org/10.18653/v1/2020.acl-main.463} {Climbing towards {NLU}: {On} meaning, form, and understanding in the age of data}.
\newblock In \emph{Proceedings of the 58th Annual Meeting of the Association for Computational Linguistics}, pages 5185--5198, Online. Association for Computational Linguistics.

\bibitem[{Brown et~al.(2020)Brown, Mann, Ryder, Subbiah, Kaplan, Dhariwal, Neelakantan, Shyam, Sastry, Askell et~al.}]{brown2020language}
Tom Brown, Benjamin Mann, Nick Ryder, Melanie Subbiah, Jared~D Kaplan, Prafulla Dhariwal, Arvind Neelakantan, Pranav Shyam, Girish Sastry, Amanda Askell, et~al. 2020.
\newblock Language models are few-shot learners.
\newblock \emph{Advances in neural information processing systems}, 33:1877--1901.

\bibitem[{Chen et~al.(2024)Chen, Han, and Miyao}]{chen2024multi}
Bowen Chen, Namgi Han, and Yusuke Miyao. 2024.
\newblock A multi-perspective analysis of memorization in large language models.
\newblock \emph{arXiv preprint arXiv:2405.11577}.

\bibitem[{Chen et~al.(2022)Chen, Liu, Meng, and Liang}]{chen2022revisiting}
Guanzheng Chen, Fangyu Liu, Zaiqiao Meng, and Shangsong Liang. 2022.
\newblock Revisiting parameter-efficient tuning: Are we really there yet?
\newblock \emph{arXiv preprint arXiv:2202.07962}.

\bibitem[{Chen et~al.(2023)Chen, Chen, Lan, Chen, and Pan}]{chen2023contributions}
Lingwei Chen, Ting Chen, Tianjiao Lan, Chu Chen, and Jay Pan. 2023.
\newblock The contributions of population distribution, healthcare resourcing, and transportation infrastructure to spatial accessibility of health care.
\newblock \emph{INQUIRY: The Journal of Health Care Organization, Provision, and Financing}, 60:00469580221146041.

\bibitem[{Chowdhery et~al.(2022)Chowdhery, Narang, Devlin, Bosma, Mishra, Roberts, Barham, Chung, Sutton, Gehrmann et~al.}]{chowdhery2022palm}
Aakanksha Chowdhery, Sharan Narang, Jacob Devlin, Maarten Bosma, Gaurav Mishra, Adam Roberts, Paul Barham, Hyung~Won Chung, Charles Sutton, Sebastian Gehrmann, et~al. 2022.
\newblock Palm: Scaling language modeling with pathways.
\newblock \emph{arXiv preprint arXiv:2204.02311}.

\bibitem[{Cosentino et~al.(2024)Cosentino, Belyaeva, Liu, Furlotte, Yang, Lee, Schenck, Patel, Cui, Schneider et~al.}]{cosentino2024towards}
Justin Cosentino, Anastasiya Belyaeva, Xin Liu, Nicholas~A Furlotte, Zhun Yang, Chace Lee, Erik Schenck, Yojan Patel, Jian Cui, Logan~Douglas Schneider, et~al. 2024.
\newblock Towards a personal health large language model.
\newblock \emph{arXiv preprint arXiv:2406.06474}.

\bibitem[{Dalvi et~al.(2022)Dalvi, Tafjord, and Clark}]{dalvi2022towards}
Bhavana Dalvi, Oyvind Tafjord, and Peter Clark. 2022.
\newblock Towards teachable reasoning systems: Using a dynamic memory of user feedback for continual system improvement.
\newblock In \emph{Proceedings of the 2022 Conference on Empirical Methods in Natural Language Processing}, pages 9465--9480.

\bibitem[{Daulton et~al.(2021)Daulton, Balandat, and Bakshy}]{daulton2021parallel}
Samuel Daulton, Maximilian Balandat, and Eytan Bakshy. 2021.
\newblock Parallel bayesian optimization of multiple noisy objectives with expected hypervolume improvement.
\newblock \emph{Advances in Neural Information Processing Systems}, 34:2187--2200.

\bibitem[{Deng et~al.(2022)Deng, Li, Zhang, Ding, and Lam}]{deng2022toward}
Yang Deng, Yaliang Li, Wenxuan Zhang, Bolin Ding, and Wai Lam. 2022.
\newblock Toward personalized answer generation in e-commerce via multi-perspective preference modeling.
\newblock \emph{ACM Transactions on Information Systems (TOIS)}, 40(4):1--28.

\bibitem[{Diao et~al.(2023)Diao, Xu, Xu, Wang, and Zhang}]{diao2023mixture}
Shizhe Diao, Tianyang Xu, Ruijia Xu, Jiawei Wang, and Tong Zhang. 2023.
\newblock Mixture-of-domain-adapters: Decoupling and injecting domain knowledge to pre-trained language models memories.
\newblock \emph{arXiv preprint arXiv:2306.05406}.

\bibitem[{Eriksson and Jankowiak(2021)}]{eriksson2021high}
David Eriksson and Martin Jankowiak. 2021.
\newblock High-dimensional bayesian optimization with sparse axis-aligned subspaces.
\newblock In \emph{Uncertainty in Artificial Intelligence}, pages 493--503. PMLR.

\bibitem[{Geva et~al.(2021)Geva, Schuster, Berant, and Levy}]{geva-etal-2021-transformer}
Mor Geva, Roei Schuster, Jonathan Berant, and Omer Levy. 2021.
\newblock \href {https://doi.org/10.18653/v1/2021.emnlp-main.446} {Transformer feed-forward layers are key-value memories}.
\newblock In \emph{Proceedings of the 2021 Conference on Empirical Methods in Natural Language Processing}, pages 5484--5495, Online and Punta Cana, Dominican Republic. Association for Computational Linguistics.

\bibitem[{Guu et~al.(2020)Guu, Lee, Tung, Pasupat, and Chang}]{guu2020retrieval}
Kelvin Guu, Kenton Lee, Zora Tung, Panupong Pasupat, and Mingwei Chang. 2020.
\newblock Retrieval augmented language model pre-training.
\newblock In \emph{International conference on machine learning}, pages 3929--3938. PMLR.

\bibitem[{He et~al.(2021)He, Zhou, Ma, Berg-Kirkpatrick, and Neubig}]{he2021towards}
Junxian He, Chunting Zhou, Xuezhe Ma, Taylor Berg-Kirkpatrick, and Graham Neubig. 2021.
\newblock Towards a unified view of parameter-efficient transfer learning.
\newblock \emph{arXiv preprint arXiv:2110.04366}.

\bibitem[{Houlsby et~al.(2019)Houlsby, Giurgiu, Jastrzebski, Morrone, De~Laroussilhe, Gesmundo, Attariyan, and Gelly}]{houlsby2019parameter}
Neil Houlsby, Andrei Giurgiu, Stanislaw Jastrzebski, Bruna Morrone, Quentin De~Laroussilhe, Andrea Gesmundo, Mona Attariyan, and Sylvain Gelly. 2019.
\newblock Parameter-efficient transfer learning for nlp.
\newblock In \emph{International Conference on Machine Learning}, pages 2790--2799. PMLR.

\bibitem[{Hu et~al.(2021)Hu, Wallis, Allen-Zhu, Li, Wang, Wang, Chen et~al.}]{hu2021lora}
Edward~J Hu, Phillip Wallis, Zeyuan Allen-Zhu, Yuanzhi Li, Shean Wang, Lu~Wang, Weizhu Chen, et~al. 2021.
\newblock Lora: Low-rank adaptation of large language models.
\newblock In \emph{International Conference on Learning Representations}.

\bibitem[{Huang et~al.(2023)Huang, Gutierrez, Kamana, and MacNeil}]{huang2023memory}
Ziheng Huang, Sebastian Gutierrez, Hemanth Kamana, and Stephen MacNeil. 2023.
\newblock Memory sandbox: Transparent and interactive memory management for conversational agents.
\newblock In \emph{Adjunct Proceedings of the 36th Annual ACM Symposium on User Interface Software and Technology}, pages 1--3.

\bibitem[{Kang et~al.(2023)Kang, Ni, Mehta, Sathiamoorthy, Hong, Chi, and Cheng}]{kang2023llms}
Wang-Cheng Kang, Jianmo Ni, Nikhil Mehta, Maheswaran Sathiamoorthy, Lichan Hong, Ed~Chi, and Derek~Zhiyuan Cheng. 2023.
\newblock Do llms understand user preferences? evaluating llms on user rating prediction.
\newblock \emph{arXiv preprint arXiv:2305.06474}.

\bibitem[{Kim et~al.(2024)Kim, Kim, and Bak}]{kim2024pema}
HyunJin Kim, Young~Jin Kim, and JinYeong Bak. 2024.
\newblock Pema: An offsite-tunable plug-in external memory adaptation for language models.
\newblock In \emph{Proceedings of the 2024 Conference of the North American Chapter of the Association for Computational Linguistics: Human Language Technologies (Volume 1: Long Papers)}, pages 6045--6064.

\bibitem[{Korbak et~al.(2023)Korbak, Shi, Chen, Bhalerao, Buckley, Phang, Bowman, and Perez}]{korbak2023pretraining}
Tomasz Korbak, Kejian Shi, Angelica Chen, Rasika~Vinayak Bhalerao, Christopher Buckley, Jason Phang, Samuel~R Bowman, and Ethan Perez. 2023.
\newblock Pretraining language models with human preferences.
\newblock In \emph{International Conference on Machine Learning}, pages 17506--17533. PMLR.

\bibitem[{Levenson and Sweatt(2005)}]{levenson2005epigenetic}
Jonathan~M Levenson and J~David Sweatt. 2005.
\newblock Epigenetic mechanisms in memory formation.
\newblock \emph{Nature Reviews Neuroscience}, 6(2):108--118.

\bibitem[{Lewis et~al.(2020)Lewis, Perez, Piktus, Petroni, Karpukhin, Goyal, K{\"u}ttler, Lewis, Yih, Rockt{\"a}schel et~al.}]{lewis2020retrieval}
Patrick Lewis, Ethan Perez, Aleksandra Piktus, Fabio Petroni, Vladimir Karpukhin, Naman Goyal, Heinrich K{\"u}ttler, Mike Lewis, Wen-tau Yih, Tim Rockt{\"a}schel, et~al. 2020.
\newblock Retrieval-augmented generation for knowledge-intensive nlp tasks.
\newblock \emph{Advances in Neural Information Processing Systems}, 33:9459--9474.

\bibitem[{Li and Liang(2021)}]{li2021prefix}
Xiang~Lisa Li and Percy Liang. 2021.
\newblock Prefix-tuning: Optimizing continuous prompts for generation.
\newblock In \emph{Proceedings of the 59th Annual Meeting of the Association for Computational Linguistics and the 11th International Joint Conference on Natural Language Processing (Volume 1: Long Papers)}, pages 4582--4597.

\bibitem[{Liu et~al.(2023)Liu, Liu, Zhou, Lv, Zhou, and Zhang}]{liu2023chatgpt}
Junling Liu, Chao Liu, Peilin Zhou, Renjie Lv, Kang Zhou, and Yan Zhang. 2023.
\newblock Is chatgpt a good recommender? a preliminary study.
\newblock \emph{arXiv preprint arXiv:2304.10149}.

\bibitem[{Liu et~al.(2024)Liu, Lin, Hewitt, Paranjape, Bevilacqua, Petroni, and Liang}]{liu2024lost}
Nelson~F Liu, Kevin Lin, John Hewitt, Ashwin Paranjape, Michele Bevilacqua, Fabio Petroni, and Percy Liang. 2024.
\newblock Lost in the middle: How language models use long contexts.
\newblock \emph{Transactions of the Association for Computational Linguistics}, 12:157--173.

\bibitem[{Liu et~al.(2019)Liu, Ott, Goyal, Du, Joshi, Chen, Levy, Lewis, Zettlemoyer, and Stoyanov}]{liu2019roberta}
Yinhan Liu, Myle Ott, Naman Goyal, Jingfei Du, Mandar Joshi, Danqi Chen, Omer Levy, Mike Lewis, Luke Zettlemoyer, and Veselin Stoyanov. 2019.
\newblock Roberta: A robustly optimized bert pretraining approach.
\newblock \emph{arXiv preprint arXiv:1907.11692}.

\bibitem[{Loshchilov and Hutter(2018)}]{loshchilov2018decoupled}
Ilya Loshchilov and Frank Hutter. 2018.
\newblock Decoupled weight decay regularization.
\newblock In \emph{International Conference on Learning Representations}.

\bibitem[{Ma et~al.(2021)Ma, Dou, Zhu, Zhong, and Wen}]{ma2021one}
Zhengyi Ma, Zhicheng Dou, Yutao Zhu, Hanxun Zhong, and Ji-Rong Wen. 2021.
\newblock One chatbot per person: Creating personalized chatbots based on implicit user profiles.
\newblock In \emph{Proceedings of the 44th international ACM SIGIR conference on research and development in information retrieval}, pages 555--564.

\bibitem[{Madaan et~al.(2022)Madaan, Tandon, Clark, and Yang}]{madaan2022memory}
Aman Madaan, Niket Tandon, Peter Clark, and Yiming Yang. 2022.
\newblock Memory-assisted prompt editing to improve gpt-3 after deployment.
\newblock In \emph{Proceedings of the 2022 Conference on Empirical Methods in Natural Language Processing}, pages 2833--2861.

\bibitem[{Madaan et~al.(2023)Madaan, Tandon, Gupta, Hallinan, Gao, Wiegreffe, Alon, Dziri, Prabhumoye, Yang et~al.}]{madaan2023self}
Aman Madaan, Niket Tandon, Prakhar Gupta, Skyler Hallinan, Luyu Gao, Sarah Wiegreffe, Uri Alon, Nouha Dziri, Shrimai Prabhumoye, Yiming Yang, et~al. 2023.
\newblock Self-refine: Iterative refinement with self-feedback.
\newblock \emph{arXiv preprint arXiv:2303.17651}.

\bibitem[{Min et~al.(2021)Min, Lewis, Zettlemoyer, and Hajishirzi}]{min2021metaicl}
Sewon Min, Mike Lewis, Luke Zettlemoyer, and Hannaneh Hajishirzi. 2021.
\newblock Metaicl: Learning to learn in context.
\newblock \emph{arXiv preprint arXiv:2110.15943}.

\bibitem[{Mu{\~n}oz et~al.(2024)Mu{\~n}oz, Yuan, and Jain}]{munoz2024shears}
J~Pablo Mu{\~n}oz, Jinjie Yuan, and Nilesh Jain. 2024.
\newblock Shears: Unstructured sparsity with neural low-rank adapter search.
\newblock \emph{arXiv preprint arXiv:2404.10934}.

\bibitem[{Munoz et~al.(2024)Munoz, Yuan, Zheng, and Jain}]{munoz2024lonas}
Juan~Pablo Munoz, Jinjie Yuan, Yi~Zheng, and Nilesh Jain. 2024.
\newblock Lonas: Elastic low-rank adapters for efficient large language models.
\newblock In \emph{Proceedings of the 2024 Joint International Conference on Computational Linguistics, Language Resources and Evaluation (LREC-COLING 2024)}, pages 10760--10776.

\bibitem[{Nadel et~al.(2012)Nadel, Hupbach, Gomez, and Newman-Smith}]{nadel2012memory}
Lynn Nadel, A~Hupbach, R~Gomez, and K~Newman-Smith. 2012.
\newblock Memory formation, consolidation and transformation.
\newblock \emph{Neuroscience \& Biobehavioral Reviews}, 36(7):1640--1645.

\bibitem[{Ning et~al.(2024)Ning, Liu, Wu, Wu, Berlowitz, Prakash, Green, O'Banion, and Xie}]{ning2024user}
Lin Ning, Luyang Liu, Jiaxing Wu, Neo Wu, Devora Berlowitz, Sushant Prakash, Bradley Green, Shawn O'Banion, and Jun Xie. 2024.
\newblock User-llm: Efficient llm contextualization with user embeddings.
\newblock \emph{arXiv preprint arXiv:2402.13598}.

\bibitem[{Ouyang et~al.(2022)Ouyang, Wu, Jiang, Almeida, Wainwright, Mishkin, Zhang, Agarwal, Slama, Ray et~al.}]{ouyang2022training}
Long Ouyang, Jeffrey Wu, Xu~Jiang, Diogo Almeida, Carroll Wainwright, Pamela Mishkin, Chong Zhang, Sandhini Agarwal, Katarina Slama, Alex Ray, et~al. 2022.
\newblock Training language models to follow instructions with human feedback.
\newblock \emph{Advances in Neural Information Processing Systems}, 35:27730--27744.

\bibitem[{Paszke et~al.(2017)Paszke, Gross, Chintala, Chanan, Yang, DeVito, Lin, Desmaison, Antiga, and Lerer}]{paszke2017automatic}
Adam Paszke, Sam Gross, Soumith Chintala, Gregory Chanan, Edward Yang, Zachary DeVito, Zeming Lin, Alban Desmaison, Luca Antiga, and Adam Lerer. 2017.
\newblock Automatic differentiation in pytorch.

\bibitem[{Petrov and Macdonald(2023)}]{petrov2023generative}
Aleksandr~V Petrov and Craig Macdonald. 2023.
\newblock Generative sequential recommendation with gptrec.
\newblock \emph{arXiv preprint arXiv:2306.11114}.

\bibitem[{Pfeiffer et~al.(2020{\natexlab{a}})Pfeiffer, R{\"u}ckl{\'e}, Poth, Kamath, Vuli{\'c}, Ruder, Cho, and Gurevych}]{pfeiffer2020AdapterHub}
Jonas Pfeiffer, Andreas R{\"u}ckl{\'e}, Clifton Poth, Aishwarya Kamath, Ivan Vuli{\'c}, Sebastian Ruder, Kyunghyun Cho, and Iryna Gurevych. 2020{\natexlab{a}}.
\newblock Adapterhub: A framework for adapting transformers.
\newblock In \emph{Proceedings of the 2020 Conference on Empirical Methods in Natural Language Processing: System Demonstrations}, pages 46--54.

\bibitem[{Pfeiffer et~al.(2020{\natexlab{b}})Pfeiffer, Vuli{\'c}, Gurevych, and Ruder}]{pfeiffer2020mad}
Jonas Pfeiffer, Ivan Vuli{\'c}, Iryna Gurevych, and Sebastian Ruder. 2020{\natexlab{b}}.
\newblock Mad-x: An adapter-based framework for multi-task cross-lingual transfer.
\newblock In \emph{Proceedings of the 2020 Conference on Empirical Methods in Natural Language Processing (EMNLP)}, pages 7654--7673.

\bibitem[{Salemi et~al.(2023)Salemi, Mysore, Bendersky, and Zamani}]{salemi2023lamp}
Alireza Salemi, Sheshera Mysore, Michael Bendersky, and Hamed Zamani. 2023.
\newblock Lamp: When large language models meet personalization.
\newblock \emph{arXiv preprint arXiv:2304.11406}.

\bibitem[{Stiennon et~al.(2020)Stiennon, Ouyang, Wu, Ziegler, Lowe, Voss, Radford, Amodei, and Christiano}]{stiennon2020learning}
Nisan Stiennon, Long Ouyang, Jeffrey Wu, Daniel Ziegler, Ryan Lowe, Chelsea Voss, Alec Radford, Dario Amodei, and Paul~F Christiano. 2020.
\newblock Learning to summarize with human feedback.
\newblock \emph{Advances in Neural Information Processing Systems}, 33:3008--3021.

\bibitem[{Tandon et~al.(2021)Tandon, Madaan, Clark, and Yang}]{tandon2021learning}
Niket Tandon, Aman Madaan, Peter Clark, and Yiming Yang. 2021.
\newblock Learning to repair: Repairing model output errors after deployment using a dynamic memory of feedback.
\newblock \emph{arXiv preprint arXiv:2112.09737}.

\bibitem[{Tay et~al.(2022)Tay, Tran, Dehghani, Ni, Bahri, Mehta, Qin, Hui, Zhao, Gupta et~al.}]{tay2022transformer}
Yi~Tay, Vinh Tran, Mostafa Dehghani, Jianmo Ni, Dara Bahri, Harsh Mehta, Zhen Qin, Kai Hui, Zhe Zhao, Jai Gupta, et~al. 2022.
\newblock Transformer memory as a differentiable search index.
\newblock \emph{Advances in Neural Information Processing Systems}, 35:21831--21843.

\bibitem[{Touvron et~al.(2023)Touvron, Lavril, Izacard, Martinet, Lachaux, Lacroix, Rozi{\`e}re, Goyal, Hambro, Azhar, Rodriguez, Joulin, Grave, and Lample}]{touvron2023llama}
Hugo Touvron, Thibaut Lavril, Gautier Izacard, Xavier Martinet, Marie-Anne Lachaux, Timoth{\'e}e Lacroix, Baptiste Rozi{\`e}re, Naman Goyal, Eric Hambro, Faisal Azhar, Aurelien Rodriguez, Armand Joulin, Edouard Grave, and Guillaume Lample. 2023.
\newblock Llama: Open and efficient foundation language models.
\newblock \emph{arXiv preprint arXiv:2302.13971}.

\bibitem[{Wang et~al.(2023)Wang, Wang, Mi, Wang, Xu, and Wong}]{wang2023chain}
Hongru Wang, Rui Wang, Fei Mi, Zezhong Wang, Ruifeng Xu, and Kam-Fai Wong. 2023.
\newblock Chain-of-thought prompting for responding to in-depth dialogue questions with llm.
\newblock \emph{arXiv preprint arXiv:2305.11792}.

\bibitem[{Wang et~al.(2020)Wang, Tang, Duan, Wei, Huang, Cao, Jiang, Zhou et~al.}]{wang2020k}
Ruize Wang, Duyu Tang, Nan Duan, Zhongyu Wei, Xuanjing Huang, Guihong Cao, Daxin Jiang, Ming Zhou, et~al. 2020.
\newblock K-adapter: Infusing knowledge into pre-trained models with adapters.
\newblock \emph{arXiv preprint arXiv:2002.01808}.

\bibitem[{Wolf et~al.(2020)Wolf, Debut, Sanh, Chaumond, Delangue, Moi, Cistac, Rault, Louf, Funtowicz, Davison, Shleifer, von Platen, Ma, Jernite, Plu, Xu, Le~Scao, Gugger, Drame, Lhoest, and Rush}]{wolf-etal-2020-transformers}
Thomas Wolf, Lysandre Debut, Victor Sanh, Julien Chaumond, Clement Delangue, Anthony Moi, Pierric Cistac, Tim Rault, Remi Louf, Morgan Funtowicz, Joe Davison, Sam Shleifer, Patrick von Platen, Clara Ma, Yacine Jernite, Julien Plu, Canwen Xu, Teven Le~Scao, Sylvain Gugger, Mariama Drame, Quentin Lhoest, and Alexander Rush. 2020.
\newblock \href {https://doi.org/10.18653/v1/2020.emnlp-demos.6} {Transformers: State-of-the-art natural language processing}.
\newblock In \emph{Proceedings of the 2020 Conference on Empirical Methods in Natural Language Processing: System Demonstrations}, pages 38--45, Online. Association for Computational Linguistics.

\bibitem[{Wu et~al.(2023)Wu, Antonova, Kan, Lepert, Zeng, Song, Bohg, Rusinkiewicz, and Funkhouser}]{wu2023tidybot}
Jimmy Wu, Rika Antonova, Adam Kan, Marion Lepert, Andy Zeng, Shuran Song, Jeannette Bohg, Szymon Rusinkiewicz, and Thomas Funkhouser. 2023.
\newblock Tidybot: Personalized robot assistance with large language models.
\newblock \emph{arXiv preprint arXiv:2305.05658}.

\bibitem[{Xu et~al.(2023)Xu, Guo, Duan, and McAuley}]{xu2023baize}
Canwen Xu, Daya Guo, Nan Duan, and Julian McAuley. 2023.
\newblock Baize: An open-source chat model with parameter-efficient tuning on self-chat data.
\newblock \emph{arXiv preprint arXiv:2304.01196}.

\bibitem[{Xu et~al.(2024)Xu, Wang, Li, Liu, Wang, Liu, and Liu}]{xu2024pluggable}
Yuzhuang Xu, Shuo Wang, Peng Li, Xuebo Liu, Xiaolong Wang, Weidong Liu, and Yang Liu. 2024.
\newblock Pluggable neural machine translation models via memory-augmented adapters.
\newblock In \emph{Proceedings of the 2024 Joint International Conference on Computational Linguistics, Language Resources and Evaluation (LREC-COLING 2024)}, pages 12794--12808.

\bibitem[{Yao et~al.(2022)Yao, Huang, Dong, Wei, Chen, and Zhang}]{yao2022kformer}
Yunzhi Yao, Shaohan Huang, Li~Dong, Furu Wei, Huajun Chen, and Ningyu Zhang. 2022.
\newblock Kformer: Knowledge injection in transformer feed-forward layers.
\newblock In \emph{CCF International Conference on Natural Language Processing and Chinese Computing}, pages 131--143. Springer.

\bibitem[{Ye et~al.(2023)Ye, Liu, Chong, Zhou, Hua, and Liu}]{ye2023qilin}
Qichen Ye, Junling Liu, Dading Chong, Peilin Zhou, Yining Hua, and Andrew Liu. 2023.
\newblock Qilin-med: Multi-stage knowledge injection advanced medical large language model.
\newblock \emph{arXiv preprint arXiv:2310.09089}.

\bibitem[{Yuan et~al.(2022)Yuan, Wang, Wang, Fine, Abdelghani, Lucas, Sauz{\'e}on, and Oudeyer}]{yuan2022selecting}
Xingdi Yuan, Tong Wang, Yen-Hsiang Wang, Emery Fine, Rania Abdelghani, Pauline Lucas, H{\'e}l{\`e}ne Sauz{\'e}on, and Pierre-Yves Oudeyer. 2022.
\newblock Selecting better samples from pre-trained llms: A case study on question generation.
\newblock \emph{arXiv preprint arXiv:2209.11000}.

\bibitem[{Zhang et~al.(2023)Zhang, Zhao, Kang, and Liu}]{zhang2023memory}
Kai Zhang, Fubang Zhao, Yangyang Kang, and Xiaozhong Liu. 2023.
\newblock Memory-augmented llm personalization with short-and long-term memory coordination.
\newblock \emph{arXiv preprint arXiv:2309.11696}.

\bibitem[{Zhang et~al.(2020)Zhang, Sun, Galley, Chen, Brockett, Gao, Gao, Liu, and Dolan}]{zhang2019dialogpt}
Yizhe Zhang, Siqi Sun, Michel Galley, Yen-Chun Chen, Chris Brockett, Xiang Gao, Jianfeng Gao, Jingjing Liu, and Bill Dolan. 2020.
\newblock Dialogpt: Large-scale generative pre-training for conversational response generation.
\newblock In \emph{ACL, system demonstration}.

\bibitem[{Zhong et~al.(2022)Zhong, Dou, Zhu, Qian, and Wen}]{zhong2022less}
Hanxun Zhong, Zhicheng Dou, Yutao Zhu, Hongjin Qian, and Ji-Rong Wen. 2022.
\newblock Less is more: Learning to refine dialogue history for personalized dialogue generation.
\newblock \emph{arXiv preprint arXiv:2204.08128}.

\bibitem[{Zhou et~al.(2023)Zhou, Wan, Vuli{\'c}, and Korhonen}]{zhou2023autopeft}
Han Zhou, Xingchen Wan, Ivan Vuli{\'c}, and Anna Korhonen. 2023.
\newblock Autopeft: Automatic configuration search for parameter-efficient fine-tuning.
\newblock \emph{arXiv preprint arXiv:2301.12132}.

\end{thebibliography}
\bibliographystyle{acl_natbib}

\appendix


\section{Case Study}
\label{appendix:case_study}
\begin{figure*}
    \centering
    \includegraphics[width=\textwidth]{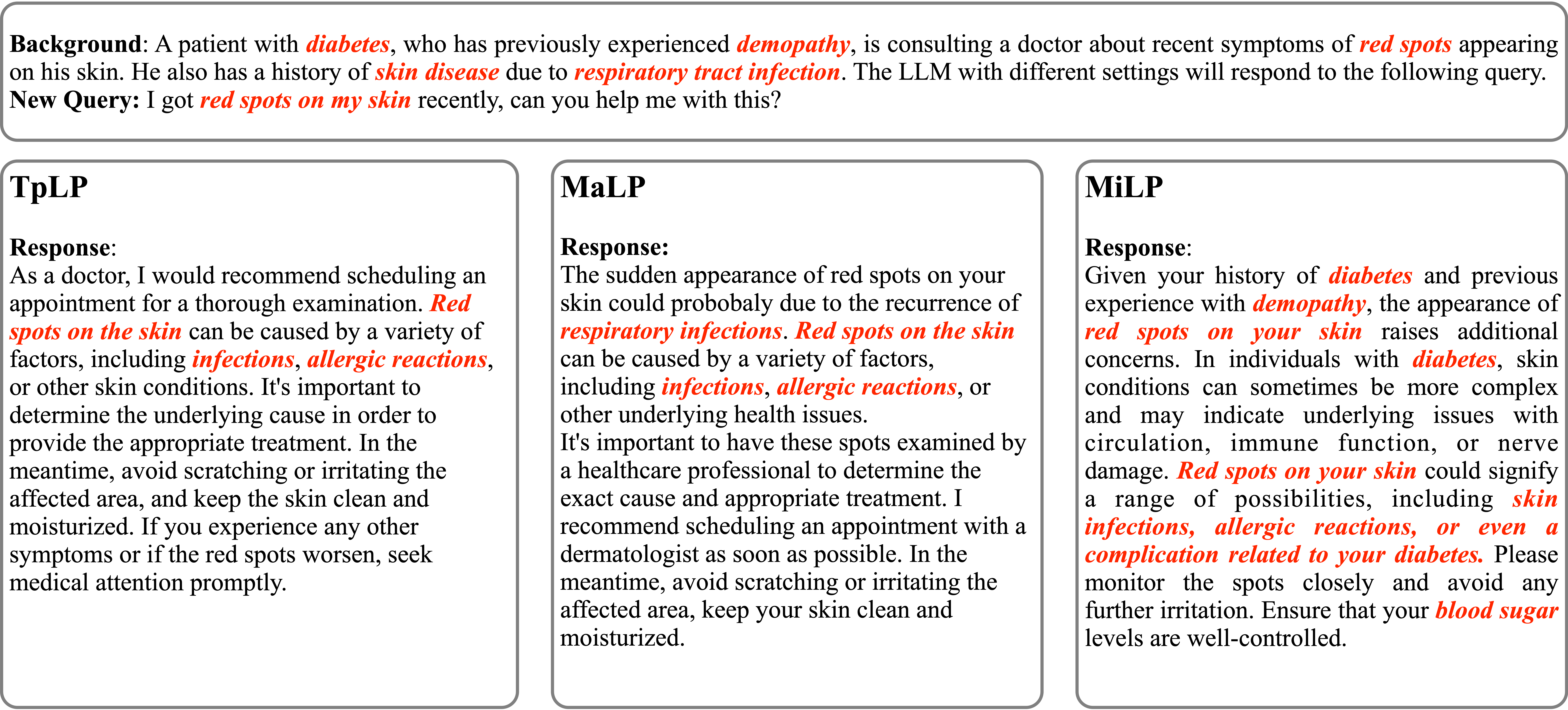}
    \caption{Case study of generated responses by LlaMA2-13B under three different settings based on provided background information.}
    \label{fig:case_study}
    \vspace{-10pt}
\end{figure*}

As can be seen in Figure \ref{fig:case_study}, Tex-prompt-based memory only perceives that this patient has \textit{skin disease} and give general reasons as well suggestions. MaLP, on the other hand, captures the \textit{respiratory infection} information due to its long- and short-term memory coordination and gives a more targeted analysis. However, the suggestions are still in general. Our MiLP, with the ability of understanding and inferring user-specific information, produces more detailed responses such as "blood sugar control", "circulation" etc. It analyzes the potential causes and gives suggestions from both diabetes and skin infection aspects. The response of MiLP covers the most personal information against other two settings which show the high quality of generated responses in terms of penalization and further confirm the power of our proposed method. 

\section{Scalability Justification}
\label{appendix:scalability}
First, the proposed MiLP offers a unique approach distinct from existing works. MiLP is not attempting to undermine the value of existing training-free/user embedding works; rather, it aims to provide a fresh perspective to the community on the benefits of parameterized memory. Secondly, we recognize that the requirements for personalization can vary across different real-world scenarios. For example, in a medical assistant context, retrieving incorrect information can lead to catastrophic outcomes (e.g., dosage recommendation etc.) in downstream tasks. In such high-stakes scenarios, relying solely on retrievers, despite their simplicity and effectiveness, may raise accuracy concerns. Therefore, it might be more beneficial to leverage the LLM’s natural language understanding and inference capabilities to utilize personal information effectively, rather than depending on similarity-based methods. We believe that our team, as well as the broader community, will continue to explore and build upon MiLP and other existing works to discover more applicable and effective methods for various scenarios.

\section{Retrieval vs Parameterized Memory}
\label{appendix:ragVSmilp}
Existing works have demonstrated that parameterized memory can be pluggable, either through a cross-attention mechanism \citep{xu2024pluggable} or using adapters\citep{kim2024pema}. This means that a user's memory can be encapsulated within a module and, when needed, loaded and integrated into a large language model (LLM) for personalization. Regarding retrievability, unlike traditional similarity-based methods, the retrieval process in parameterized memory functions by activating relevant parameters in response to a given query, resulting in a personalized output. When a new user is introduced, we can load the corresponding parameterized module into the base model to generate personalized responses. However, if there are no pre-existing user records, our method requires training. Our team is dedicated to integrating Retrieval-Augmented Generation (RAG) with parameterized memory modules. This integration allows new users without prior records to provide preliminary information to the LLM using the RAG method. Over time, as interactions continue, our MiLP can parameterize these interactions for more accurate and personalized responses.

\end{document}